\documentclass[pdflatex,apa]{sn-jnl}
\usepackage{graphicx}%
\usepackage{multirow}%
\usepackage{amsmath,amssymb,amsfonts}%
\usepackage{amsthm}%
\usepackage{mathrsfs}%
\usepackage[title]{appendix}%
\usepackage{xcolor}%
\usepackage{textcomp}%
\usepackage{manyfoot}%
\usepackage{booktabs}%
\usepackage{algorithm}%
\usepackage{algorithmicx}%
\usepackage{algpseudocode}%
\usepackage{listings}%
\usepackage{comment}
\usepackage[numbers]{natbib}

\theoremstyle{thmstyleone}%

\begin{document}

\title[Article Title]{Artificial Spacetimes for Reactive Control of Resource-Limited Robots}
\author[1]{\fnm{William} \sur{H. Reinhardt}}\email{whr@seas.upenn.edu}

\author*[1]{\fnm{Marc Z.} \sur{Miskin}}\email{mmiskin@seas.upenn.edu}

\affil[1]{\orgdiv{Department of Electrical and Systems Engineering}, \orgname{University of Pennsylvania}, \orgaddress{\city{Philadelphia}, \postcode{19104}, \state{PA}, \country{USA}}}


\abstract{Field-based reactive control provides a minimalist, decentralized route to guiding robots that lack onboard computation.  Such schemes are well suited to resource-limited machines like microrobots, yet implementation artifacts, limited behaviors, and the frequent lack of formal guarantees blunt adoption.  Here, we address these challenges with a new geometric approach called artificial spacetimes.  We show that reactive robots navigating control fields obey the same dynamics as light rays in general relativity. This surprising connection allows us to adopt techniques from relativity and optics for constructing and analyzing control fields.  When implemented, artificial spacetimes guide robots around structured environments, simultaneously avoiding boundaries and executing tasks like rallying or sorting, even when the field itself is static.  We augment these capabilities with formal tools for analyzing what robots will do and provide experimental validation with silicon-based microrobots.  Combined, this work provides a new framework for generating composed robot behaviors with minimal overhead.

}

\maketitle
\newpage
\section{Introduction}\label{sec1}
Microrobots hold significant promise for a variety of applications including drug-delivery \cite{nelson_microrobots_2010,iacovacci_medical_2024}, environmental remediation \cite{zarei_self-propelled_2018,urso_smart_2023}, and micro-manufacturing \cite{zhang_robotic_2019}. However, their small size imposes significant limitations for onboard computation, sensing, and actuation, making many traditional control architectures difficult to implement.  As a result, new and unique  strategies are required to guide robots too small to see by eye. 

The majority of existing microrobot control schemes fall into two groups: top-down and reactive.  In top-down control, an external system continuously monitors each robot and directly updates forcing fields that guide it toward a desired state. These approaches, often implemented using optical tweezers \cite{zhang_distributed_2020}, electromagnetic actuation \cite{diller_independent_2013, ghosh_controlled_2009,frutiger_small_2010,khalil_closed-loop_2013}, and/or acoustic fields \cite{ahmed_selectively_2015,ahmed_artificial_2017,del_campo_fonseca_ultrasound_2023}, provide precise and adaptable control, making them well-suited for complex, multi-step tasks or those requiring high accuracy. However, existing implementations struggle with accommodating large numbers of independent microrobots, as many of the forcing fields are hard to spatiotemporally pattern and face trade-offs between resolution and field of view when scaled to large workspaces \cite{jiang_control_2022,chowdhury_controlling_2015}.  Further compounding these challenges is the fact that top-down approaches often require extensive calibration or bespoke pieces of laboratory equipment to implement.

Reactive control strategies provide a complementary approach that could help address these shortcomings, emphasizing parallelism and decentralization. Rather than relying on continuous external feedback, they use on-robot sensory inputs to immediately modulate the robot’s actions in response to a global control field.  Common examples are stimuli-responsive micromotors that  achieve taxis \cite{you_intelligent_2018}, artificial potential fields that coordinate motion through attractive and repulsive forces \cite{kim_electric_2016}, and microrobot swarms whose behavior emerges through collective interactions \cite{ceron_programmable_2023,ji_collective_2023,gardi_microrobot_2022,wang_coordinated_2020}.  Because each robot executes actions without waiting for a central controller, these approaches reduce implementation overhead and enable swarms of microrobots to operate in parallel without individualized tracking and intervention.  Furthermore, many microrobots are made from  field-responsive engines  \cite{leonardo_biohybrid_2023,hanson_electrokinetic_2024,you_intelligent_2018}, making  reactive control a natural framework.  

A major challenge for reactive control is building fields that promote sophisticated actions.  While a variety of work has demonstrated simple behaviors (e.g., taxis) \cite{you_intelligent_2018,zhuang_ph-taxis_2015,gao_micro-nano_2022}, it remains difficult to realize complex tasks like navigating a structured environment or independently guiding multiple agents to distinct places using a single, static field.  Further, many algorithms for reactive control are heuristic, lacking formal guarantees and suffering from implementation artifacts (e.g., local minima) unless further optimized \cite{rafai_review_2022,rehman_motion_2023,duhe_contributions_2021}. In cases where  control fields do yield desired results, it is challenging to port the solution to a new problem, often requiring fields to  be redesigned to match each application.  

In this work, we propose a new approach to creating reactive control fields, called artificial spacetimes, which lends itself to addressing some of these issues.  Specifically, this geometric framework  allows us rationally construct fields that Braitenberg-like robots \cite{hotton_open_2024} can use to carry out  tasks in structured environments, offers  formal methods for analyzing what a robot will do and when, and enables solutions to be reused across problems. 

Our starting point is the observation that the trajectories of certain robots under reactive control in slowly-varying fields are formally equivalent to null geodesics on a Lorentzian manifold \cite{carroll_spacetime_2004,leonhardt_transformation_2008}.  From this point of view, control fields set up local metric tensors while reactive robots follow paths of minimal space-time interval. This connection, while surprising, enables us to leverage specialized coordinate transformations and well-studied metrics from optics and relativity to solve various control problems.   For instance, we present metrics that enable robots to rally, confine, or turn in prescribed ways.  Further, we are able to embed boundary avoidance within the same control field without sacrificing the desired behavior.  In other words, these results can be composed with one another, enabling fields that simultaneously prevent robot collisions with walls and give instructions for navigation, localization, and/or sorting without requiring explicit on-robot computation. In addition to theoretical results, we validate this framework through both simulations and experiments with microrobots.  Combined, these results show a promising new way to extract interesting behaviors from small robots that lack the ability to compute.

\section{Results}\label{sec2}

In Figure \ref{fig1}a, we outline the kinematics for a common robot design that maps its local sensory inputs to motor speed (commonly known as a Braitenberg vehicle \cite{hotton_open_2024}). The robot has two motors that each move at their own velocity depending on the sensory intensity $I(x)$ (e.g., light intensity).  The body of the robot traces a path in two dimensions with curvature $\kappa$, proportional to the normalized difference in motor velocities, at a speed  $V_B$ equal to the average speed of the two motors.  When placed in a spatially varying sensor field, visualized in Figure \ref{fig1}a, the robot turns due to the unequal stimulation of the motors. We note that despite having no onboard computer or memory system, a robot navigating in this way remains under closed-loop control; the robot takes local measurements of its environment  and its speed and direction are modulated in response.

With an eye towards microrobots, we consider the case that the sensory field varies slowly in space relative to the size of the robot.  In other words, variation in $I$ takes place over a much larger length scale than  the robot's size.  Under this assumption, $I$ can be approximated locally by a Taylor series, leading to  equations (see SI) for the robot's acceleration (equation \ref{accel}) and speed  (equation \ref{bodyspeed}):
\begin{equation}\label{accel}
    I\frac{d^2x_i}{ds^2} = \epsilon_{ij}\frac{dx_j}{ds}(\frac{dx_l}{ds}\epsilon_{lm}\partial_mI)
\end{equation}
\begin{equation}\label{bodyspeed}
    \frac{ds}{dt} = I(x_i)
\end{equation}
where $s$ is the arclength traveled by the robot, $x$ is its coordinate position, and $\epsilon_{ij}$ is a $90^\circ$ counter-clockwise rotation matrix.

Both equations are strongly reminiscent of Lorentzian geodesics. First,  equation \ref{accel} is structurally consistent with the general form of a geodesic: the right hand side is quadratic in the coordinate velocities while the left hand side is an acceleration. Second,  equation \ref{bodyspeed} can be rewritten as 
   $ I^2(x,y)dt^2-dx^2-dy^2 = 0$ by squaring both sides.  
 This form shows the robots next positions are constrained to a cone of accessible spatiotemporal locations, a hallmark of Lorentzian geometry. The cone's opening angle varies in space and time, controlled by the sensory field $I$, and, as visualized in Figure \ref{fig1}b, the robot's trajectory is defined by where these cones meet.  Indeed, prior work \cite{li_robophysical_2023,li_field-mediated_2022} has shown surprising experimental connections between differential drive macro-scale robots and relativistic physics, suggesting the possibility of a formal relationship. 

Together, these observations strongly suggest that the robot's motion could be derived by minimizing an action on an appropriately curved spacetime.   By applying  calculus of variations to the  metric implied by  equation \ref{bodyspeed}, we find this the correspondence is exact:  the robot's motion is formally identical to the path light takes in a curved spacetime (see SI for derivation).

Connecting the path of a reactive control robot to a Lorentzian geodesic brings significant formal structure.  One can use known symmetries in the spacetime to predict how long it will take a robot to traverse a path or predict whether a region of space is asymptotically attracting.  Crucially, the numerous symmetries, intuitions, and solved problems from both relativistic mechanics and ray optics can be appropriated for the task of engineering robot behaviors. 

As an example, we analyze a proportional control field intended to guide robots to the origin  from the point of view of  Lorentzian geodesics.  Specifically,  we take the sensory field $I=\sqrt{x^2+y^2}=r$ such that each motor's velocity is proportional to its distance from the target location.  Heuristically, this scheme aims to both align robots to the target location and slow them down as they approach.  

Using our geometric formalism, we determine the range of initial headings and positions that are guaranteed to arrive at the origin and their rate of convergence.    Since this metric depends only on the radial distance, the dynamics conserve both energy $E=-r^2 dt/d\lambda $ and angular momentum $L=r^2 d\phi/d\lambda$ where $\lambda$ is an affine parametrization and $\phi$ is the angle between the robot's position and the origin.  Using these symmetries and the null-affine parametrization $d\lambda=I^2 dt$ \cite{visser_efficient_2023},   equation \ref{bodyspeed} simplifies to a  first order equation for the radial velocity (see SI for details)
\begin{equation}\label{approach}
(\frac{dr}{dt})^2 = r^2 (1-L^2)
\end{equation}
Since the sign of the right hand side of eqn. \ref{approach} cannot pass through zero, robots cannot change whether they are headed towards or away from the origin.  Indeed, explicitly computing the angular momentum gives $L=\sin(\alpha_0)$ where $\alpha_0$ is the initial misalignment between the robot's heading and its radial vector (see SI for details); the phase space is always split between strictly ingoing and outgoing paths.  

Focusing on the inward falling trajectories, we directly compute the rate of convergence.  Integrating eqn. \ref{approach}, we find $r(t) = r_0\exp[-t \cos(\alpha_0)]$.  In other words, any robot with an initial misalignment between $+\pi/2 $ and $-\pi/2$ converges exponentially fast to the origin, regardless of its initial distance from the target location.  Figure \ref{fig1}c and Figure \ref{fig1}d show simulations of robots compared to the theory, validating the result. No fit parameters are used.

This simple example hints at how Lorentizian geometry can  address  control problems.  Yet, the use of symmetry arguments raises the question of how to handle more realistic geometries.  A structured environment like a maze would clearly lack the radial symmetry of an unobstructed position control problem.  

This issue can be resolved by leveraging the coordinate invariance of geodesic motion and light-like trajectories \cite{post_formal_1997, leonhardt_transformation_2008}. By properly choosing a new set of coordinates, complex boundaries can be mapped into simple ones, extending the reach of rudimentary control fields. One particularly powerful class of transformations are those that locally preserve angles, or conformal transformations.  By definition, a conformal transformation from  $x \rightarrow \tilde{x}(x)$ transforms the spacetime interval as $ d\tilde{s}^2 = ds^2\Omega^2(x)$.
Because the overall effect on the interval is multiplication by scalar function,  a null geodesic in one coordinate system remains a null geodesic in the transformed space. The value arises from the fact that many known conformal transformations map complicated boundaries (e.g., arbitrary polygons) to trivial ones (e.g., a disk).  Of note, conformal mapping of geodesics is used extensively in optics to build devices ranging from lenses to mirrors to cloaks \cite{leonhardt_transformation_2008,leonhardt_optical_2006,xu_conformal_2015,turpin_conformal_2010} and transformations can often be generated with computationally efficient algorithms (e.g., root-exponential convergence to solutions ) \cite{trefethen_numerical_2020,driscoll_schwarz-christoffel_2002,driscoll_algorithm_1996}.


This invariance property presents a two-stage composition procedure for building reactive control fields, illustrated schematically in Figure \ref{fig2}. First, encode all the boundary information via conformal transformation to a simple virtual space.  In virtual space, implement the desired  behavior with a control metric.  The  conformal transformation can then be inverted to map the control law back to physical space,  resulting in a composite metric  that expresses  both the desired behavior and stores information about how to avoid obstacles. 

As proof of this approach, we simulate  robots, governed by equations \ref{accel} and \ref{bodyspeed}, navigating non-trivial boundaries under reactive control. We first demonstrate the task of maze traversal by mapping the  maze to a rectangular virtual space, imposing the proportional control metric, and inverting the transform to build the control field for physical space (see SI). As seen in Figure \ref{fig3}a, three robots with different initial positions trace out the maze and reach the  target location.  While virtual space shows essentially the same behavior as trajectories in Figure \ref{fig1}c, guiding robots to the goal, the mapping back to physical space adds the missing information needed for  robot to avoid collisions with the walls. 

This two-stage approach can be used with other virtual space metrics to complete tasks besides position control. For instance, metrics, originally developed for optics, exist to  rally, confine, rotate, or deflect trajectories  \cite{xiong_designing_2013,gallina_transformation_2008,nazarzadeh_wideband_2022,liang_wide-angle_2013, lee_optical_2021,sun_transformation_2017,alitalo_electromagnetic_2009}.   The paths in Figure \ref{fig1}a and Figure \ref{fig1}b are characteristic, showing robots patrolling a circular region \cite{tyc_absolute_2011} and focusing them to patrol along a linear path \cite{gomez-reino_design_2008,bouchard_grin_2014},  respectively.  Applied in virtual space, these same functionalities can be composed with wall avoidance to create more sophisticated behaviors.  For instance, Figure \ref{fig3}b shows a metric that turns robots by a fixed amount based on their incident heading in virtual space.  Mapped to real space, robots get sorted upon exiting a junction, each arriving at a unique position despite the fact that they use the same static control field. 

As a final test of our framework, we present experimental studies on microscale robots, showing their applicability in the physical world.  We fabricate (see SI) silicon-based microrobots (Figure \ref{fig4}a) that use electrokinetic motors in solution to move around their environments \cite{hanson_electrokinetic_2024}. As in the model, each robot has two motors, made up of discrete arrays of silicon photovoltaics, that move at speeds proportional to incident light intensity.  We utilize the optical setup in Figure \ref{fig4}a to project spatially varying intensity fields on the bottom surface of petri dish that holds the robot (see SI), enabling us to establish  tunable metrics. 


We find  agreement between the measured  robot trajectories and light-like geodesics predicted by our theory. Figure \ref{fig4}b shows several control fields corresponding to oscillatory motion along a linear axis \cite{hecht_optics_2017,gomez-reino_design_2008,bouchard_grin_2014}, orbiting trajectories \cite{tyc_absolute_2011}, and fixed-angle turning \cite{zhao_analytic_2020,jen_yung_li_90_nodate}.  Comparing  to simulated trajectories (Figure \ref{fig4}b, bottom), we find  agreement. For example, the 90$^{\circ}$ turning metric \cite{zhao_analytic_2020,jen_yung_li_90_nodate}  rotates our robot's heading $\approx 82^\circ$, with the robot path falling on a simulated geodesic. Further, as seen in the simulation, the behavior is relatively insensitive to the initial heading and provides approximate $90^\circ$ turns for a $40^\circ$ window of entry angles, suggesting this strategy could be deployed without precise knowledge of the robot's initial orientation. 

Metrics can also be directly joined in space, provided they either extrapolate to a constant value or share a boundary with the same overall intensity field values.  To demonstrate, we place two 90$^{\circ}$ turning lenses next to each other (Fig. \ref{fig4}b). Similar to optics where lenses can added or removed from the beam path to change the overall behavior, the adjacent control gradients yield a composite effect, turning the robot by a full 180$^{\circ}$. In addition to being fairly insensitive to the input angle, control metrics can accept a range of input positions while retaining their functionality. For example, in Figure \ref{fig4}b, robots with initial positions spanning approximately $33\%$ of the lens width are rotated and output with a much narrower spread ($\sim15\%$ of the width).

\section{Discussion}\label{sec5}
Artificial spacetimes provide a versatile framework for controlling robots that lack the capacity to compute.   Long term, our control strategy could potentially enable microrobots to traverse  anatomical structures for drug delivery or cover specific areas in environmental remediation applications without explicit knowledge of the robot's location.   Such results could even find use for microrobots with  on-board computation \cite{xu_210x340x50_2022,reynolds_microscopic_2022}, reducing memory requirements by offloading the information required for  navigation into a sensory field.

While here we have focused on a specific robot class (i.e., Braitenberg vehicles in 2D), there are several promising paths towards generalization.   One  route could extend the metrics to vary in time.  Many  applications of autonomous robots require multiple, conditioned, sequential steps.  As the manifold  here is fundamentally spatiotemporal, it may be possible to encode some of these actions as temporal variations in $I$ without losing geodesic structure.   Work along this route might target  robot-to-robot collision avoidance  or sequential exploration of space by causing individual robots to speed up or slow down upon arriving at specific spacetime locations.  As an added benefit,  temporally varying metrics are of interest in related scientific fields, namely optical metamaterials \cite{engheta_four-dimensional_2023,leonhardt_transformation_2009} and experimental analog gravity \cite{barcelo_analogue_2011}.  As dynamic metrics are generally difficult to realize, robots have a unique opportunity to guide scientific explorations by analogy.

A second generalization could expand the robot hardware.  While here we use silicon microrobotics with electrokinetic motors, many microrobot platforms have a mapping between propulsion speed and local measurements like chemical concentration \cite{feng_advances_2023}, light intensity \cite{leonardo_biohybrid_2023}, or temperature \cite{ji_thermoresponsive_2019}. Provided there is a way to create a spatially varying control field, our two-stage composition procedure can be adopted. Alternatively,  the governing geometric principles can be generalized to 3D: in the supporting material we specify robot dynamics that extend equations \ref{accel} and \ref{bodyspeed} to produce Lorentzian geodesic motion in three spatial dimensions.  

A final generalization could allow  robots to establish the control field themselves.  While here we project the governing metric onto the system, other  works have explored how swarms of reactive  robots can realize emergent behaviors by creating their own control fields or interactions \cite{ceron_programmable_2023,defay_characterization_2022,gauci_self-organized_2014}.  In this case, the geometry couples to what the robots are doing, in effect  making the metric a dynamical variable and moving even closer in spirit to general relativity.  While the metric dynamics would likely be different, relativistic mathematical and numerical tools might lend themselves to analyzing and controlling the collectively formed geometry, motion, and behaviors.

\newpage
\begin{figure}[H]
\centering
\includegraphics[width=12cm]{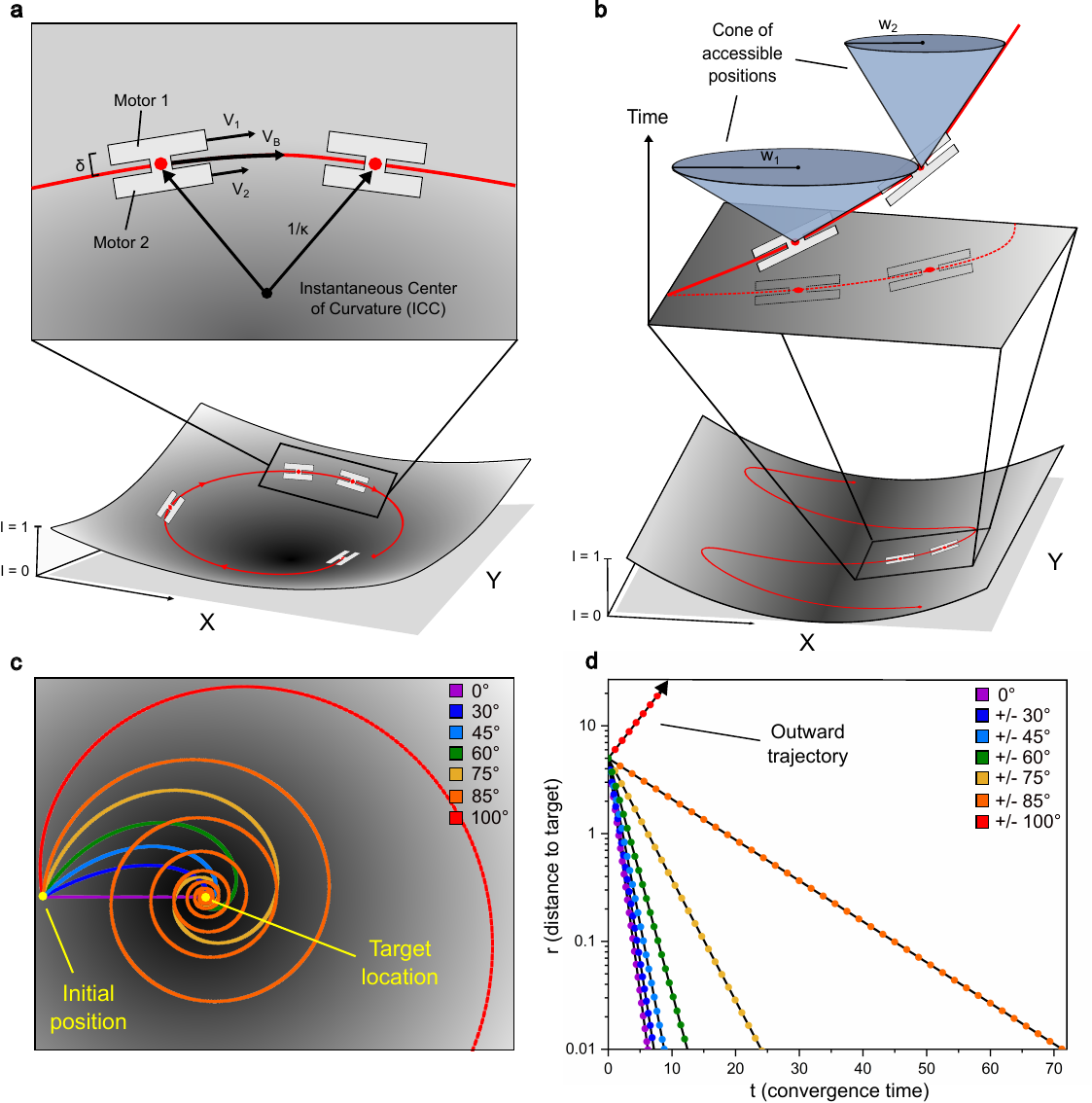}
\caption{\textbf{Motion in gradient fields. a)} On top, the kinematics of a basic robot with two motors that move at speeds $V_1$ and $V_2$. The overall body speed $V_B$ is given by the average of the motor velocities, and the curvature $\kappa$ is proportional to the normalized difference in motor velocity. The robot's width is given by $2\delta$. Below, a plot of normalized intensity illustrates the control field in three dimensions. \textbf{b)} Plotted with the third dimension of time, robots travel along trajectories connecting cones of accessible positions. The width of the cone $w_1$ or $w_2$ depends on the velocity of the robot. \textbf{c)} Trajectories with an initial angle in the range $+\pi/2 \rightarrow -\pi/2$ are guaranteed to reach the target location, and trajectories outside of this range escape. \textbf{d)} Convergence time for a variety of initial orientations. The black lines are the predicted results for the derived exponential relationship $r(t) = r_0exp(-t\cos(\alpha_0))$.}
\label{fig1}
\end{figure}

\begin{figure}[H]
\centering
\includegraphics[width=12cm]{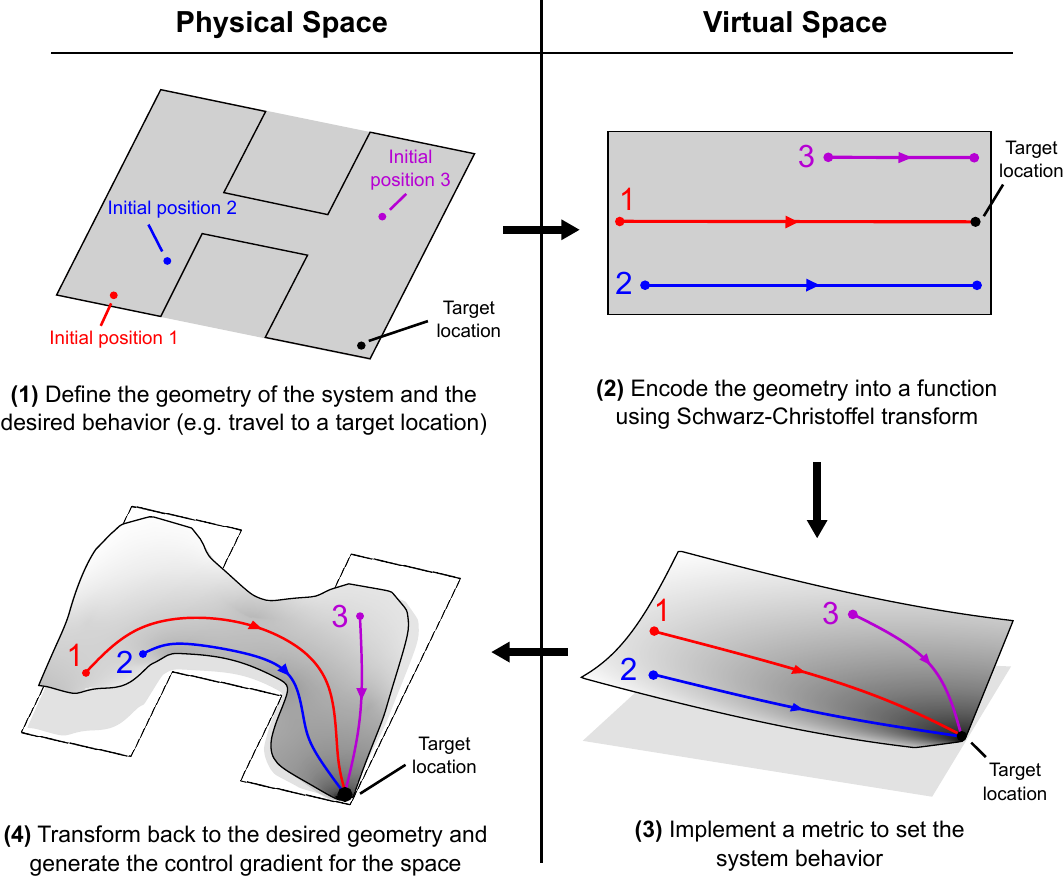}
\caption{\textbf{Two-stage process to building artificial spacetimes.} Coordinate transformations enable metrics for behaviors to be composed with metrics for boundary avoidance.  First, we map a physical space with a complex boundary to a simple virtual space by a conformal transformation.  A control metrics is then implemented in virtual space to dictate the robot's behavior. For example, here robots all travel to the target location. When mapped back to physical space, we produce a composite metric that both encodes the behavior and boundary information in a single scalar field.}
\label{fig2}
\end{figure}

\begin{figure}[H]
\centering
\includegraphics[width=12cm]{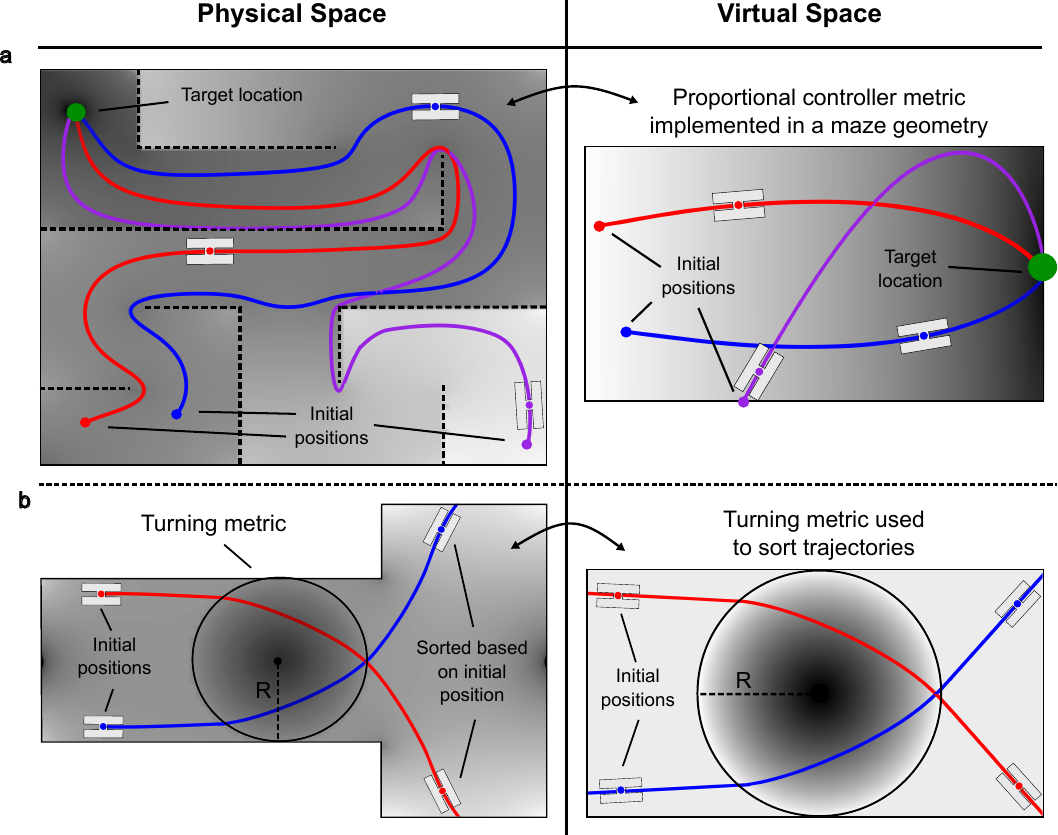}
\caption{\textbf{Simulating robot trajectories. a)} A maze traversal demonstration with an embedded proportional control metric that directs robots to the same target location. For visualization,  the control field magnitude is plotted  logarithmically. \textbf{b)} A turning metric acts as a sorter that directs robots to different sides of the space based on the initial trajectory.  Note both robots use the same static field.  }
\label{fig3}
\end{figure}

\begin{figure}[H]
\centering
\includegraphics[width=12cm]{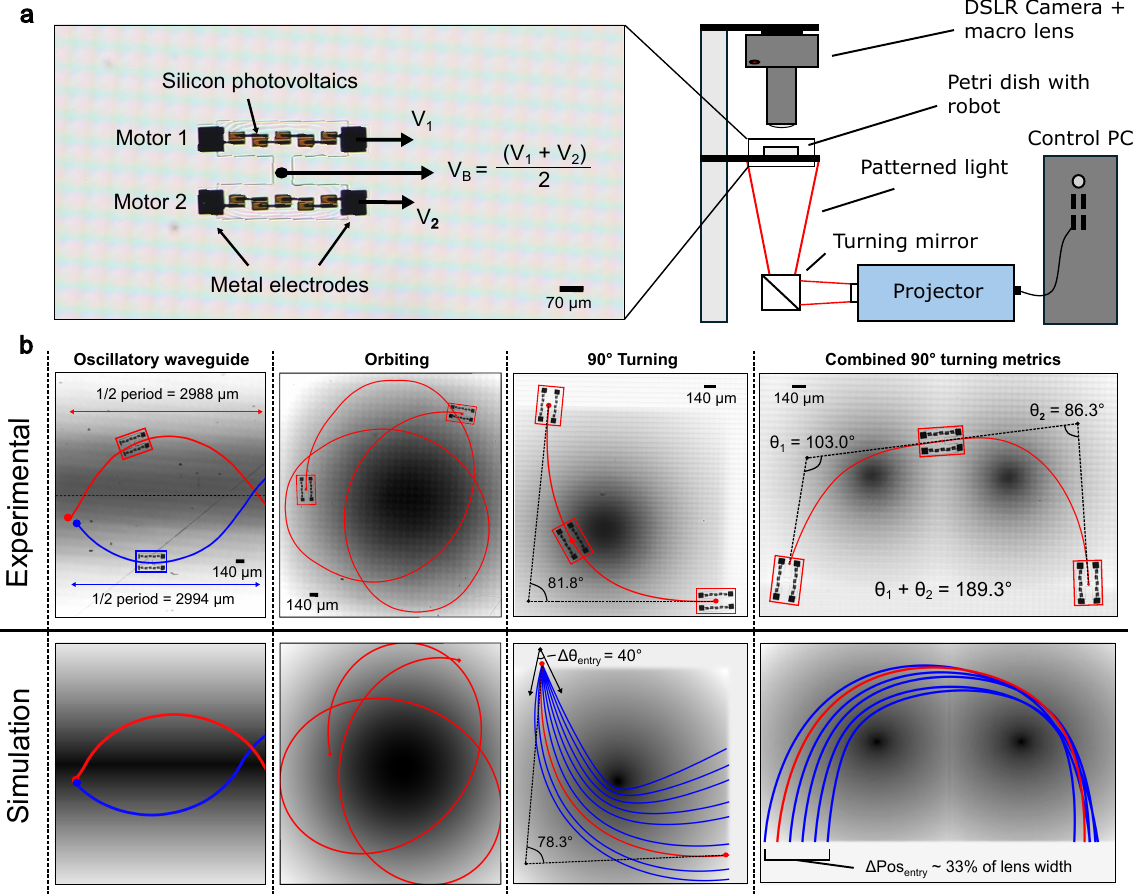}
\caption{\textbf{Experimental demonstrations of the general relativity correspondence. a)} A silicon-based microrobot with two electrokinetic motors and differential drive kinematics and an optical setup that creates spatially varying intensity fields. \textbf{b)}  Experimental demonstrations of various control metrics implemented as intensity fields (top) compared to simulation results (bottom). In the 90$^\circ$ and double $90^\circ$ turning metric experiments, the simulated traces in red most closely match the initial pose of the experimental demonstration above.}
\label{fig4}
\end{figure}

\backmatter

\bmhead{Supplementary information}
\bmhead{Deriving Equations of Motion}
We derive the equations of motion for a robot with the kinematics shown in Figure \ref{fig1}a traveling through a spatially varying sensory field. Akin to a differential drive robot, each motor moves at a speed $V_1$ or $V_2$. The speed along a path is given as 
\begin{equation}\label{diffdrive}
    \frac{ds}{dt} = \frac{1}{2}(V_1 + V_2)
\end{equation}
and the rate of change of the instantaneous center of curvature (ICC) vector is
\begin{equation}\label{ICCvec}
    \frac{d\phi}{dt} = \frac{(V_1-V_2)}{w}
\end{equation}
defining $w$ as the distance from the center of one motor to the center of the other. We can rewrite equation \ref{ICCvec} using the robot's heading vector $\chi = [\cos(\theta),\sin(\theta)]$ where $\theta$ is the heading angle with respect to the global coordinate system. By definition of the ICC, $d\theta = -d\phi$, so using the chain rule we get
\begin{equation}
    \frac{d\chi_i}{dt} = -[-\sin(\phi),\cos(\phi)]\frac{d\theta}{dt}
\end{equation}
or equivalently,
\begin{equation}
    \frac{d\chi_i}{dt} = \epsilon_{ij}\chi_j\frac{d\phi}{dt}
\end{equation}
where $\epsilon_{ij}$ is a 90$^\circ$ counter-clockwise rotation matrix. Substituting in our definition of $\frac{d\phi}{dt}$, we get a set of motion equations in terms of the motor velocities.
\begin{equation}\label{speedeq}
    \frac{ds}{dt} = \frac{1}{2}(V_1 + V_2)
\end{equation}
\begin{equation}\label{roteq}
    \frac{d\chi_i}{dt} = \epsilon_{ij}\chi_j\frac{(V_1-V_2)}{w}
\end{equation}
Next, we aim to define these equations in terms of the intensity $I$. If we assume that the variation in motor speeds comes from a spatial gradient in intensity that is small over the robot's body, we can approximate intensity at a point as 
\begin{equation}
    I(x,y) \approx I(x_0,y_0) + \partial_iI\Delta x_i + \Delta x_i\partial_{ij}I\Delta x_j
\end{equation}
To find the average intensity over a motor, we choose $\Delta x_i = \epsilon_{ij}\chi_j(\frac{w}{2}) + \chi_iu$ where $u$ spans the length of the robot body, i.e. $u = [-l/2,l/2]$. We then integrate over $u$ to obtain the average intensity on the motor.
\begin{equation}
    I_1 \approx I_0 + \frac{w}{2}\partial_iI\epsilon_{ij}\chi_j + \frac{w^2}{8}\epsilon_{il}\chi_l\partial_{ij}I(\epsilon_{jk}\chi_k)+\frac{l^2}{24}\chi_i\partial_{ij}I\chi_j
\end{equation}
The intensity over the other motor can be found by setting $w = -w$.
\begin{equation}
    I_2 \approx I_0 - \frac{w}{2}\partial_iI\epsilon_{ij}\chi_j + \frac{w^2}{8}\epsilon_{il}\chi_l\partial_{ij}I(\epsilon_{jk}\chi_k)+\frac{l^2}{24}\chi_i\partial_{ij}I\chi_j
\end{equation}
Assuming light intensity is proportional to motor speed, we can substitute $I_1$ and $I_2$ for $V_1$ and $V_2$ in equation \ref{speedeq} and equation \ref{roteq} to get our equations of motion in terms of the intensity,

\begin{equation}
    \frac{ds}{dt} = I + \frac{w^2}{8}(\epsilon_{ij}\chi_l)[\partial_{ij}I](\epsilon_{jk}\chi_k) + \frac{l^2}{24}\chi_i[\partial_{ij}I]\chi_j
\end{equation}

\begin{equation}
    \frac{d\chi_i}{dt} = \epsilon_{ij}\chi_j(\chi_l\epsilon_{lm}\partial_mI)
\end{equation}

In the limit that $w,l \rightarrow 0$ and using $\chi_i = \frac{dx_i}{ds}$, the final equations of motion are
\begin{equation}
    \frac{ds}{dt} = I(x,y)
\end{equation}
\begin{equation}
    I\frac{d^2x_i}{ds^2} = \epsilon_{ij}\frac{dx_j}{ds}(\frac{dx_l}{ds}\epsilon_{lm}\partial_mI(x,y))
\end{equation}

\bmhead{Finding Geodesic Equations}
We follow the standard procedure for deriving the geodesic equation by varying the action. Light-like trajectories satisfy the null geodesic condition $g_{\mu\nu}dx^\mu dx^\nu = 0$ where $g$ is the system's metric tensor and $dx^\mu = [dt,dx,dy]$ \cite{carroll_spacetime_2004}. Using $ I^2(x,y)dt^2-dx^2-dy^2 = 0$ from squaring the speed equation, we get a metric
\begin{equation}
    g \sim \begin{pmatrix}
    I^2(x,y) & 0\\
    0 & -\delta^{ij}
\end{pmatrix}
\end{equation}
We can write the action in terms of the line element $ds$, 
\begin{equation}
    S = \int ds^2
\end{equation}
To get the geodesic equation, we vary this action with respect to an affine parameterization of the curve, $\lambda$,
\begin{equation}
    S = \int g_{\mu\nu}\frac{dx^\mu}{d\lambda}\frac{dx^\nu}{d\lambda}d\lambda
\end{equation}
For the given metric $g$, we choose a parameterization $d\lambda \propto I^2dt = Ids$ \cite{visser_efficient_2023}. Varying the action with respect to $x$,
\begin{equation}
    \frac{d}{d\lambda}[-2\frac{dx}{d\lambda}] = 2I(x,y)\frac{dI}{dx}\frac{dt^2}{d\lambda}
\end{equation}
Using our parameterization of $\lambda$, we can substitute $\frac{dt}{d\lambda} = \frac{1}{I^2}$ and $d\lambda = Ids$ to get
\begin{equation}
    \frac{1}{I}\frac{d}{ds}[\frac{1}{I}\frac{dx}{ds}] = -\frac{dI}{dx}\frac{1}{I^3}
\end{equation}
We apply the chain rule and get
\begin{equation}
    \frac{1}{I^2}[\frac{d^2x}{ds^2}]-\frac{1}{I^3}(\frac{dI}{dx}\frac{dx}{ds} + \frac{dI}{dy}\frac{dy}{ds})\frac{dx}{ds} = -\frac{dI}{dx}\frac{1}{I^3} 
\end{equation}
Substituting $1 = (\frac{dx}{ds})^2 + (\frac{dy}{ds})^2$ on the right hand side,
\begin{equation}
    \frac{d^2x}{ds^2}-\frac{1}{I}(\frac{dI}{dx}(\frac{dx}{ds})^2 + \frac{dI}{dy}\frac{dy}{ds}\frac{dx}{ds}) = -\frac{dI}{dx}\frac{1}{I}((\frac{dx}{ds})^2 + (\frac{dy}{ds})^2)
\end{equation}
We can then cancel $-\frac{1}{I}(\frac{dI}{dx}(\frac{dx}{ds})^2)$ from both sides, leaving us with 
\begin{equation}
    \frac{d^2x}{ds^2} = -\frac{dI}{dx}\frac{1}{I}(\frac{dy}{ds})^2 + \frac{1}{I}\frac{dI}{dy}\frac{dy}{ds}\frac{dx}{ds}
\end{equation}
Rearranging, we have the light-like geodesic equation for this space,
\begin{equation}
    I\frac{d^2x}{ds^2} = -\frac{dy}{ds}(-\frac{dx}{ds}\frac{dI}{dy} + \frac{dy}{ds}\frac{dI}{dx})
\end{equation}

Using the relation $\chi_i = \frac{dx_i}{ds}$, we can rewrite this as

\begin{equation}
    I\frac{d^2x}{ds^2} = -\chi_y(-\chi_x\partial_yI+\chi_y\partial_xI)
\end{equation}

The $y$ component follows a similar derivation, leading to the general equation

\begin{equation}
    I\frac{d^2x_i}{ds^2} = \epsilon_{ij}\frac{dx_j}{ds}(\frac{dx_l}{ds}\epsilon_{lm}\partial_mI)
\end{equation}
where $\epsilon$ is a $90^\circ$ counter-clockwise rotation matrix. This is identical to the robot's equation of motion, equation \ref{accel}, derived from differential drive kinematics.

\bmhead{Analysis of the Proportional Metric}
Here we provide a detailed derivation of the region and rate of convergence for the proportional control metric, whose spacetime interval is, in polar coordinates:

\begin{equation}
ds^2 =-r^2 dt^2 + dr^2 +r^2d\phi^2
\end{equation}

We can note immediately that the metric does not have any explicit dependence on $t$ or on $\phi$.  Thus, the dynamics admit two conserved quantities corresponding to the generalized momentum for these coordinates.  Explicitly, if we write the action principle as 

\begin{equation}
S=\int d \lambda \mathcal{L}=\int d\lambda \frac{1}{2}[-r^2 \frac{dt}{d\lambda}^2 + \frac{dr}{d\lambda}^2 +r^2\frac{d\phi}{d\lambda}^2]
\end{equation}

Then the momenta $E=\partial_{dt/d\lambda} \mathcal{L}=-r^2dt/d\lambda$ and $L=\partial_{d\phi/d\lambda} \mathcal{L}=r^2d\phi/d\lambda$ are constants of motion.  

Rather than explicitly solve the geodesics, we can insert these conserved quantities into the null condition for our geodesics.  Such an analysis is standard in relativity, and used to determine closest approach in a  two-body problem \cite{carroll_spacetime_2004}.

Specifically, inserting the constants of motion to reduce $-r^2 \frac{dt}{d\lambda}^2 + \frac{dr}{d\lambda}^2 +r^2\frac{d\phi}{d\lambda}^2=0$ gives us 

\begin{equation}
0 =-\frac{E^2}{r^2} + \frac{dr}{d\lambda}^2 +\frac{L^2}{r^2}
\end{equation}

Finally, we use the known null-affine parametrization for this metric form \cite{visser_efficient_2023}, $d\lambda=I^2dt=r^2dt$ to arrive at eqn. \ref{approach}:

\begin{equation}
 (\frac{dr}{dt})^2 =r^2(E^2-L^2)
\end{equation}

The final steps in our derivation are linking $L$ and $E$ back to robot parameters for comparison.  In our explicit affine parametrization, $E=-r^2dt/(r^2dt)=-1$ for all initial placements.  For the angular part, we find $L=d\phi/dt$.  To simplify comparison, we can rewrite this result using the heading vector and initial velocity as 

\begin{equation}
\frac{d\phi}{dt}=\frac{1}{r^2}[x\frac{dy}{dt}-y\frac{dx}{dt}]]
\end{equation}
Or noting that $\frac{1}{r} \frac{d\vec{x}}{dt} $ is the normalized heading of the robot $\vec{\chi}$ while $\vec{x}/r=\hat{r}$ we can write 

\begin{equation}
L=\hat{r}\times\vec{\chi}=\sin[\alpha_0] 
\end{equation}
where the last line follows as $\hat{r}$ and $\vec{\chi}$ are both unit length.  

With the above expressions for $L$ and $E$ we can  derive the rate of convergence.  Explicitly, eqn. \ref{approach}, becomes:
\begin{equation}
 (\frac{dr}{dt})^2 =r^2(1-\sin[\alpha_0]^2)=r^2\cos[\alpha_0]^2
\end{equation}

We again note that the right hand side of the expression can only reach zero when the robot halts at the origin.  Thus, any in-falling trajectory is always in-falling while any out-falling trajectory is always out-falling.  Since the sign of $dr/dt$ is fixed from the initial conditions, we can take the square root of both sides and apply the appropriate sign based on  initial orientation.  The result is simple first order differential equation for the radial coordinate which is easily solved to give the convergence equation from the main text:

\begin{equation}
 r(t)=r_0 \exp[-\cos[\alpha_0]t]
\end{equation}

As a final comment, we point out that the single-signed nature of convergence is unique to this metric: most others feature a so-called centrifugal barrier in which the conserved angular momentum sets a distance of closest approach.  By selecting a proportional metric, we eliminate this problem since the $E$ and $L$ terms in the governing radial equation follow the same functional dependence on R.  This  eliminates the possibility of a special distance at which $dr/dt$ vanishes, and by extension, precludes turning points.

\bmhead{Generating Intensity Profiles}
Recent work in transformation optics has demonstrated the capability to guide and control light by building spatially varying index of refraction profiles. Because our robot's motion is light-like, we can implement these same techniques for robot control. Of note, in this work, there is an inverse relationship between index of refraction and intensity. Light travels slower in a medium with a higher index of refraction, and the robots discussed here travel faster at higher intensity. However, this is simply handled by taking the index of refraction profile $n$ used to guide light and calculating $\frac{1}{n}$ to build an intensity profile.

Many transformation optics components, such as spherically-symmetric lenses and GRIN lenses, have straightforward functions corresponding to their index of refraction profiles \cite{tyc_absolute_2011}. To generate these components as an intensity field, we calculate the profile over a grid of points, take the inverse, and map these values to a grayscale image (0-255). Various components are generated in this way and implemented in Figure \ref{fig4}.

Further, it is well-known in transformation optics how to build an index of refraction profile corresponding to a conformal mapping function $w$ \cite{leonhardt_optical_2006,xu_conformal_2015,turpin_conformal_2010}. Simply taking the magnitude of the map's derivative will produce a profile that embeds the coordinate transform, and arbitrary metrics such as proportional controllers can be encoded by multiplying a metric $n_{virtual}$ with this profile, i.e.
\begin{equation}
    n_{physical} = n_{virtual}|\frac{dw}{dz}|
\end{equation}
Again, keeping in mind the inverse relationship between $n$ and $I$, it is straightforward to build an intensity profile that handles both the geometry of the system and the control law.

\bmhead{Simulating Robot Trajectories}
Robot trajectories are simulated using Python's 'solve\_ivp' differential equation solver. To start, we manually initialize the robot's position and heading inside the desired geometry in physical space. Then, we locate the motor positions by rotating the heading vector $90^\circ$ and incrementing by $\delta$ (half our simulated robot width) in each direction. Once we have the motor locations, we map these points to virtual space and calculate the intensity on each motor by taking the magnitude of the mapping function derivative and it multiplying with the virtual space metric. As intensity is directly proportional to velocity, we can calculate $\frac{dx}{dt}$ and $\frac{dy}{dt}$ by multiplying with the relevant component of the heading vector
\begin{equation}
    \frac{dx}{dt} = \cos(\theta)(I_{left}+I_{right})/2
\end{equation}
\begin{equation}
    \frac{dy}{dt} = \sin(\theta)(I_{left}+I_{right})/2
\end{equation}
It is also straightforward to calculate the heading change,
\begin{equation}
    \frac{d\theta}{dt} = \frac{(I_{left}-I_{right})}{2\delta}
\end{equation}
At each timestep, these values are calculated and passed to the solver to initialize the next iteration. Notably, this entire simulation takes place in physical space, only utilizing virtual space to look up the corresponding intensity value.

\bmhead{Microrobot Fabrication}
Robots are fabricated massively in parallel using standard semiconductor fabrication tools and techniques as previously demonstrated by our group \cite{hanson_electrokinetic_2024,miskin_electronically_2020}. To start, we take silicon on insulator (SOI) wafers with a 2 micron (p-type) device layer on top of 500 nm insulating silicon dioxide. These two layers sit on 500 microns of handle silicon. 
We fabricate photovoltaics by diffusing n-type dopants into the silicon device layer using phosphorus-based spin on glass (Filmtronics P509) in a rapid thermal annealer. We plasma etch mesa structures to expose the underlying p-type silicon, conformally deposit silicon dioxide to insulate the entire sample, and form contacts to the p-type and n-type regions by selectively HF etching the oxide and then sputtering titanium and platinum into the holes.
Next, we form interconnects between the photovoltaics and the electrodes that act as the robot's actuator by sputtering titanium and platinum. We encapsulate the entire structure in SU8 for insulation, plasma etch the silicon dioxide layer to define the robot body, and release robots from the SOI by covering the sample in aluminum and underetching the handle silicon with XeF2 vapor. We then dissolve the aluminum support film to free robots into solution.

\bmhead{Experimental Optical Setup}
To form spatially varying intensity fields, we use a commercially available DLP projector (NP-V332W) connected to a computer and display grayscale images corresponding to the control fields. Utilizing the projector's direct mapping between grayscale pixel value (0-255) and the light intensity, projected fields manifest as spatially varying intensity fields on the bottom surface of the petri dish containing our robots. For imaging, we use a DSLR camera (Canon Mark IV EOS) with a macro lens (Laowa 25mm f/2.8 2.5-5X Ultra Macro).

Due to the limitations our optical system and finite turning rate for our robots, we are constrained balancing  tasks complexity against image size. We find that in our current projector setup, the smallest optical features we can make are approximately 40x40 um, the size of a single photovoltaic. This constrains the resolution of our intensity profiles, preventing (spatially) small changes in the gradient from contributing effectively to the turning kinematics. For example, near walls and sharp turns, the gradient changes rapidly-- an effect which may not be accurately captured in our experimental setup due to the resolution. 
Further, the robot itself has a finite radius of curvature due to the electrokinetic actuation scheme that prevents sharp, pivot turns even when there is a large difference in intensity on each motor. 

\bmhead{Transformation Optics Components as Control Fields}
In this work, we utilize a GRIN lens and two spherically-symmetric lenses called the Maxwell fisheye and the Eaton lens. Each of these has a well-defined index of refraction profile utilized in transformation optics. As such, we can directly generate these components as intensity fields for the robot by utilizing the inverse relationship between index of refraction and intensity.

The GRIN lens has a central optical axis with an intensity minimum that spans the length of the lens according to the optical profile
\begin{equation}
    n(s) = n_0(1-\frac{As^2}{2})
\end{equation}
where $n_0$ is the baseline index of refraction value (e.g. 1), $A$ is a constant known as the gradient parameter, and $s$ is the distance from the center optical axis. Light oscillates sinusoidally as it travels through the lens with a consistent period.

The first of the spherically symmetric lenses, the Maxwell fisheye, has an optical profile corresponding to the following equation \cite{tyc_absolute_2011}
\begin{equation}
    n(r) = \frac{2}{1+r^2}
\end{equation}
where $r$ is the distance from the center point of the lens. Light-like trajectories in these profiles circle the center.

The second spherically symmetric lens is the Eaton lens. It has an approximate profile \cite{kim_retroreflector_2012} given by
$n(r) \approx (\frac{2}{r}-1)^{\frac{\theta}{\theta+\pi}}$
where $r$ is the distance to the center of the lens and theta is the desired turning angle.

\bmhead{Extending Results to 3D}

While in this work we have focused on robots confined to moving in a plane, many of the underlying concepts can  be translated to 3D.  As an explicit example, consider a robot that moves with a speed fixed by its position
\begin{equation}\label{3dbody}
\frac{ds}{dt}=I(x,y,z)
\end{equation}
along a heading vector $\vec{\chi}$.  Treating $\vec{\chi}$ as a three dimensional vector, we seek a generalization of equation \ref{accel} that preserves the magnitude of $\vec{\chi}$ under time evolution, depends on the gradient in $I$, and acts to turn the robot away from regions of high intensity.  These conditions can all be satisfied by choosing:

\begin{equation}
\frac{d\vec{\chi}}{dt}=-(\vec{\chi}\times \nabla I)\times \vec{\chi}
\end{equation}

Heuristically, this equation acts to rotate away from regions where I is large while the cross product ensures the magnitude of $\vec{\chi}$ is clearly fixed.  To show it represents a good generalization of equation \ref{accel} we can consider what happens when the intensity gradients and heading are purely co-planar.  In this case,  the term $\vec{\chi}\times \nabla I=\hat{z}(\chi_x\partial_yI-\chi_y \partial_xI)=\hat{z}\chi_i \epsilon_{ij}\partial_j I$ while the second cross product reduces to $-\hat{z}\times\vec{\chi}=\epsilon_{ij}\chi_j$.  In other words, when restricted to motion in 2D, we recover eqns. \ref{bodyspeed} and \ref{accel}, as anticipated.

To connect this 3D motion to relativistic geodesics in a 3+1 spacetime, we cast the dynamics into a slightly different form by applying the double cross product identity

\begin{equation}\label{3dacc}
\frac{d}{dt}{\vec{\chi}}(t)
\;=\;
-\nabla I
\;+\;
\Bigl(\nabla I \cdot \vec{\chi}(t)\Bigr)\,\vec{\chi}(t).
\end{equation}

This result makes it clear that the motion of the robot is to descend gradients in the sensory field, subject to the constraint that it preserves the length of the heading.  
We then compare this result to  geodesic equations resulting from treating eqn. \ref{3dbody} as an interval, giving the action principle

\begin{equation}
S=\int d\lambda [(\frac{dt}{d\lambda})^2I^2(x,y,z)-(\frac{d\vec{x}}{d\lambda})^2]
\end{equation}

Varying with respect to the spatial coordinates gives 

\begin{equation}
\frac{d}{d\lambda}[-\frac{d x_i}{d\lambda }  ]= I\partial_i I (\frac{dt}{d\lambda})^2
\end{equation}

Again, using the affine parametrization \cite{visser_efficient_2023} $d\lambda = I^2 dt$ we find

\begin{equation}
\frac{d}{dt}[-\frac{d x_i}{dt } \frac{1}{I^2} ]= \frac{\partial_i I }{I}
\end{equation}

We now note that $\frac{dx}{dt}/I=\chi_x$, letting us factor terms on the LHS to yield

\begin{equation}
\frac{d}{dt}[-\chi_x \frac{1}{I} ]= \frac{\partial_i I }{I}
\end{equation}

After applying the chain rule and re-arranging we find

\begin{equation}
\frac{d\chi_x}{dt}= -{\partial_i I }+\chi_x (\partial_l I \chi_l)
\end{equation}

The same derivation applies for all spatial components.  When we merge these results by explicitly writing them in vector form, we reproduce  the robot's governing equation for its heading vector (i.e. eqn \ref{3dacc}):

\begin{equation}
\frac{d\vec{\chi}}{dt}= -{\nabla I }+\vec\chi (\nabla I \cdot \vec\chi)
\end{equation}

Thus, the simple dynamics of a robot following equations \ref{3dacc} and \ref{3dbody} both generalize the 2D motion studied here and are formally identical to light rays in a fully 3+1 dimensional spacetime.

\bmhead{Acknowledgements}
The authors would also like to thank Professor Nader Engheta, Caroline Zhao, and Jordan Shegog for helpful discussions.  This work was supported by the Army Research Office (ARO YIP W911NF-17-S-0002), the Sloan Foundation, the Packard Foundation, and was carried out at the Singh Center for Nanotechnology, which is supported by the NSF National Nanotechnology Coordinated Infrastructure Program under grant NNCI-2025608.

\section*{Declarations}
\begin{itemize}
\item Funding:
Army Research Office (ARO YIP W911NF-17-S-0002), the Sloan Foundation, the Packard Foundation
\item Competing interests:
The authors declare no competing interests.
\item Ethics approval and consent to participate:
Not applicable
\item Consent for publication: The authors consent to publish under the CC-BY Open Access License.
\item Data availability: Experimental data is provided as supplemental videos, but raw image files are available upon request.
\item Materials availability:
Not applicable
\item Code availability:
Code to simulate robot trajectories will be hosted on Github (https://github.com/whreinhardt/).
\item Author contribution:
W.H.R and M.Z.M designed the research. W.H.R fabricated the robots, wrote the simulation, performed experiments, and analyzed data. M.Z.M conceived of the original idea, developed the theory, and supervised research. W.H.R and M.Z.M wrote the paper.
\end{itemize}

\bibliographystyle{unsrtnat}
\bibliography{references.bib}

\end{document}